%% file: paper.tex
\documentclass{article}
\usepackage{arxiv}
\usepackage{framed,multirow}

\usepackage{amssymb}
\usepackage{latexsym}

\usepackage{url}
\usepackage{amssymb}
\usepackage[cmex10]{amsmath}
\usepackage{boldline,booktabs}
\usepackage{varwidth}
\usepackage{enumitem}
\usepackage[dvipsnames]{xcolor}
\usepackage{tikz}
\usetikzlibrary{fadings,shapes.arrows,arrows,shapes,fit,positioning,calc,shadows,matrix,decorations.pathreplacing}
\usepackage{pgfplots}
\pgfplotsset{compat=newest}
\definecolor{FireBrick}{rgb}{0.7, 0.13, 0.13}

\graphicspath{{images/}}

\tikzfading[name=arrowfading, top color=transparent!0, bottom color=transparent!95]
\tikzset{arrowfill/.style={top color=CornflowerBlue!20, bottom color=Cerulean, general shadow={fill=black, shadow yshift=-0.4ex, path fading=arrowfading}}}
\tikzset{arrowstyle/.style={draw=FireBrick,arrowfill, single arrow,minimum height=#1, single arrow,
single arrow head extend=.2cm,}}

\newcommand{\tikzfancyarrow}[2][2cm]{\tikz[baseline=-0.5ex]\node [arrowstyle=#1] {#2};}

\tikzset{
    position/.style args={#1:#2 from #3}{
        at=(#3.#1), anchor=#1+180, shift=(#1:#2)
    }
    }

\newcommand{\commentOriol}[1]{{\bf \color{red} ORIOL: #1}}
\newcommand{\commentBhaskar}[1]{{\bf \color{blue} Bhaskar: #1}}

\newcommand{\EA}{\text{EA}}

\title{Knowledge graph based methods for record linkage}

\date{June 2019}

\author{
  B. Gautam, O. Ramos Terrades\\
  Computer Vision Center - Dept. of Computer Science\\
  Universitat Aut\`{o}noma de Barcelona\\
  Cerdanyola del Vallés, 08193 \\
  \texttt{{oriolrt}@cvc.uab.cat} 
  \And %
    J. M. Pujades, M. Valls  \\
  Demographic Center Studies \\
  Universitat Aut\`{o}noma de Barcelona,\\
   Edifici E2 Campus UAB, \\
   Bellaterra 08193, Spain
}

\begin{document}
\maketitle
\begin{abstract}
Nowadays, it is common in Historical Demography the use of individual-level data as a consequence of a predominant life-course approach for the understanding of the demographic behaviour, family transition, mobility, etc. Record linkage advance is key in these disciplines since it allows to increase the volume and the data complexity to be analyzed. However, current methods are constrained to link data coming from the same kind of sources. Knowledge graph are flexible semantic representations, which allow to encode data variability and semantic relations in a structured manner.

In this paper we propose the knowledge graph use to tackle record linkage task. The proposed method, named {\bf WERL}, takes advantage of the main knowledge graph properties and learns embedding vectors to encode census information. These embeddings are properly weighted to maximize the record linkage performance. We have evaluated this method on benchmark data sets and we have compared it to related methods with stimulating and satisfactory  results. 
\end{abstract}

%



\section{Introduction}

Nowadays, it is common in Historical Demography the use of individual-level data as a consequence of a predominant life-course approach for the understanding of the demographic behaviour, family transition, mobility, etc. The building of individual life-courses involves often the link of the birth certificate (baptism), the death certificate of the same individual and, for those who got married, their marriage certificate. These certificates were compiled nominally and manually in the so called Quinque Libri (parish registers) from the Council of Trento in the 16th century onwards throughout Europe~\cite{Willigan2013}. Moreover and as consequence of the establishment of the Liberal State in Europe, civil registers and censuses spread out from the 19th century, although in some countries spread out even before~\cite{Garcia2007,Garcia2012,Valero1986}. Consequently, individual life-courses can also be enriched by sequential observations of the same individual over time and space using census material. 

In addition to these two information sources, other studies in the discipline apply an inter-generational perspective such as the inter-generational transmission of demographic behaviour, social outcomes or diseases, which is key in genetic studies and  necessarily requires the connection across generations~\cite{Fu2014}. Moreover, combining data from different sources, which at the same time involves a cross-validate data, help create a larger understanding of individual demographic and social behaviour. Thus, individual life-courses, which are built from parish records, are enriched by sequential observations of the same individual over time and space using census material. Indeed, {\bf record linkage} is the task of finding records in a data set that refer to the same entity across different data sources in which individuals do not have a unique identifier~\cite{Ruggles:2017}.

Record linkage advance is key for disciplines like Historical Demography since it allows to increase the volume and the data complexity to be analyzed. Early development in computing devices like the Hollerith Tabulator  was a result of increasing census data size in late 17th century and the lack of resources to tabulate and to analyze this information~\cite{Kistermann2005}. Since then, and mostly from the 19th century on-wards, many states  gathered population censuses with the purpose of rationalizing the state population and wealth~\cite{Yule1919,woolf1989,Porter1995,Raphael2008}. State statistics institutionalisation  was a response to the wish of quantifiable material within an epistemological development framework  of scientific objectiveness and  an impersonal knowledge of issues and phenomena~\cite{Porter1995}. Hence, current government research departments uses have extensively used their census data for population projection and for provision plan in local governments. Moreover, it is worth noting how various countries have made their historical census data publicly available  (see IPUMS NAPP\footnote{\url{https://www.nappdata.org/napp/}} and Mosaic project\footnote{\url{https://censusmosaic.demog.berkeley.edu/}}).

However, poor data quality can turn record linkage into a challenging task. The conservation status of  original documents, the scanning process, the large number of similar values in names, ages or addresses are just some of the multiple factors that can affect data quality. More importantly, the relationship between  household members and household head can  significantly change between two record censuses. For example, people were born and died, got married, changed occupation, or moved home. These changes on individual's records make record linkage in  early historical census, i.e. those collected in the 19th and early 20th century and where only limited information about individuals were available, even more challenging. As a result, record linkage methods are not reliable enough, and many false or duplicate matches are often generated. This is also a common problem in other record linkage applications, such as author disambiguation~\cite{Elmagarmid2007}.

Knowledge graphs (KG), which is also called linked data in terms of semantic web, allow organizing semantics in a structured manner. They represent a collection of interlinked descriptions of real-world objects, events, situations or abstract concepts, which are called entities, in a formal structure. Popular KG include Freebase, DBpedia, YAGO, Satori, etc. and they have millions of entities and billions of entity links. Many researchers have developed various link prediction methods using KG embedding techniques since these large KG  are still incomplete with  some missing links~\cite{Bordes2013, Wang2014,Lin2015}.

These link prediction methods, like record linkage methods, are also known as entity alignment (EA) methods. EA methods aims at aligning entities between different entity groups or even different KGs~\cite{Guan2017}. 
A few studies have employed KG  embeddings to conduct EA that neither require to define similarity measures nor to make assumptions in advance.
Actually and as an important property of KGs, entities contain rich semantic information that can be utilized to improve EA  methods performance~\cite{Hao2016}.

Therefore, there is the need of record linkage methods able to deal with heterogeneous data sources. Existing methods are tailored to a particular set of attributes and this makes more difficult record linkage across heterogeneous sources. Furthermore, these methods must be able to cope with data variability, due to individual life-course changes and  robust to acquisition errors and unverified transcriptions as well. KGs provide a conceptual framework in which record linkage prediction  and cross-linkage, between different sources, are naturally defined. Indeed, the proposed method, named WERL, is a step forward in record linkage methods since it overcomes some of the main difficulties discussed earlier. The main contributions of this paper are:%
\begin{itemize}
    \item {\bf Evolution knowledge graph}. We use KGs for census data and we enrich them  to take into account changes on attribute values of linked records. 
    \item A two step learning process. Thereby, we  separately learn attribute embeddings and  their weight in record linkage tasks. This allow higher flexibility on embeddings learning and a faster training.
\end{itemize}

In summary, we propose an EA method able to deal with heterogeneous data and that can be applied to other type of KGs in which entities are described in terms of attributes. 

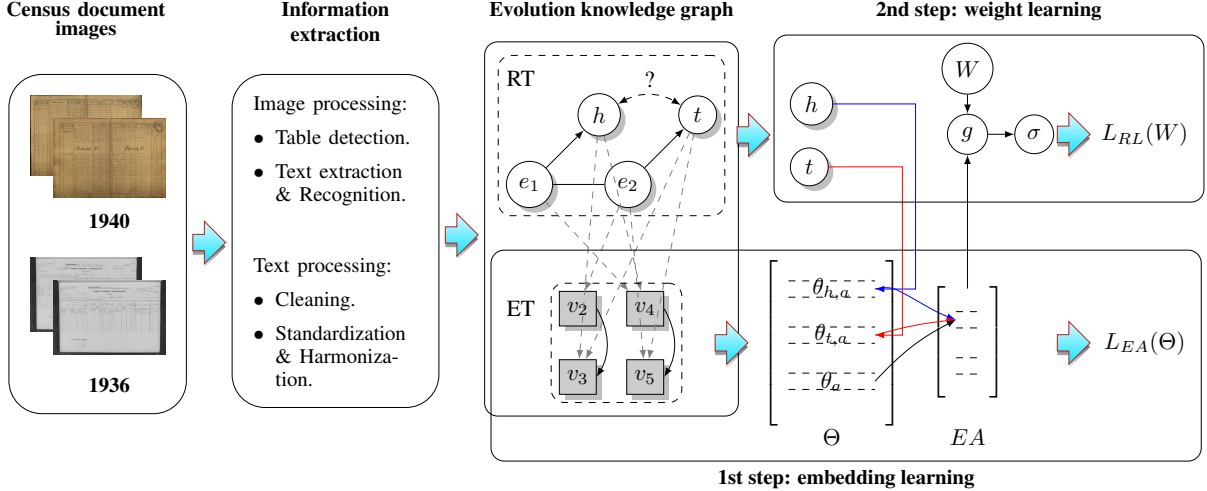
\begin{figure*}[ht]
\centering
\resizebox{\textwidth}{!}{
\input{pipeline.tex}
}
\caption{WERL overview. Document image analysis techniques and information extraction methods are first applied. Then, a KG representation with evolution triples is built. Learning is a two step process: first, attribute and attribute values are learned. Second, consist of attribute weights learning.}
\label{fig:method}
\end{figure*}

\section{Related work}
\label{sec:soa}

In this section we briefly review the main methods related to the record linkage task. First, we summarize the main methods which are devoted to this task. Then, we review those KG based methods which are closely related to the proposed method.

\subsection{Record linkage methods}

The  record linkage task has been explored by a number of disciplines, including databases, statistics and artificial intelligence. Each discipline has formulated the problem slightly different and consequently different techniques have been proposed~\cite{Ravikumar2004}. In the database community, this task is also known  as deduplication. Deduplication aims at eliminating repeated data or multiple copies and thus compressing the database. To this end, it has been proposed various string edit-distance based methods for record matching like a general-purpose scheme~\cite{Monge1997}  or a knowledge-intensive approach~\cite{Raman2001}. 

In statistics, a long line of research has been conducted in probabilistic record linkage, largely based on the seminal paper by~\cite{Fellegi1969}. The authors in \cite{Fellegi1969} formulate entity matching as a classification problem, where the basic goal is to classify entity pairs as matching or non-matching. They  propose to use unsupervised methods, based on a feature-based representation of pairs which are manually designed and are, to some extent, problem-specific. Although this can be a major problem when linking data from different sources, these proposals have been, by and large, adopted by subsequent researchers, often with elaborations of the underlying statistical model. The Jaro-Winkler distance has been used for record linkage purposes~\cite{Jaro1995,Winkler1999}. 

The AI community has focused on applying supervised learning to the record-linkage task for parameter learning of string-edit distance metrics~\cite{Ristad1998} and combining the results of different distance functions~\cite{Tejada2001}. Unsupervised methods include similarity-based ones \cite{Nikolov2012}, probabilistic ones~\cite{Febian2011}, hierarchical graph model-based ones~\cite{Ravikumar2004} and self-learning and embedding based entity alignment~\cite{Guan2017}.

\subsection{Knowledge-based representations}\label{sec:kg_methods}

The most basic knowledge representation is the entity-relation (ER) graph, whose nodes are entities and the adjacency matrix represent the entity relations. The Translations in the embedding space (TransE) method  generates knowledge graph embeddings of entities and relations, so that two related entities must have close entity embeddings in the embedding space~\cite{Bordes2013}. A major drawback of this model is the poor accuracy while modelling one-to-many and many-to-many relations.

In order to overcome this problem, the authors in \cite{Wang2014} proposed the Translating on hyperplanes (TransH) method. In that work, an entity is first projected onto a hyperplane given for each relation. Thus, the method projects the entity according to the relation it is involved in. This allows the method to perform well even with one-to-many and many-to-many relations. Moreover, in \cite{Lin2015} the authors observe that the optimal dimension used to embed entities and relations could be different since they are completely different objects. In order to address this issue they proposed the TransR method, which learns a  matrix to project embedded entities into the relation embedding space. This method performs well with one-to-many and many-to-many relations since it uses a different projection matrix for each relation.

RESCAL  is a compositional method, which generates relational embeddings using tensor factorization~\cite{Nickel2011}. Each embedding dimension represent a single entity feature, thus if two entities have the same feature value, both should have similar values in the same embedding dimension. Similarly, the Holographic Embedding (HOLE) method combines the expressive power of the tensor product with the efficiency and simplicity of TransE by using the circular correlation of vectors to represent pairs of entities~\cite{Nickel2016}. In addition, Path-based TransE (PTransE)  is an extension of TransE method, which considers multi-step relation paths along with direct links~\cite{Lin2015PTransE}. Finally, the  Bootstrapping Entity Alignment (BootEA) method  aims to update entity embedding using bootstrap process which adds likely alignments to the knowledge graph~\cite{Sun2018}.

The KR-EAR method relies in a entity-attribute-relation (EAR) knowledge representation~\cite{Lin2016}. It that work, the authors identity two kind of elements in their knowledge graph: entities, which are related between them by means of relations, and attributes, which are a sort of entity features. Consequently, they define the EA task as a product of two posterior probabilities. One modeling entities and relations and the other modeling attributes and entities. Moreover, self-learning and active learning techniques are used for embedding learning when not enough annotated data is available. The Self-Learning and Embedding based entity alignment (SEEA) method is an extension of the KR-EAR method~\cite{Guan2017}. There, the authors applies reinforcement learning techniques to improve the performance of their method.

\section{The WERL method}

The {\bf Weighted embedding based record linkage} (WERL) method relies in an extension of the EAR graph and its parameters are learned in a two step process. The full training process is depicted in~\figurename~\ref{fig:method}. First of all, there are the usual document image analysis processes consisting of layout analysis, text line detection and text recognition. Then, the extracted text have to be processed, cleaned, standardized and harmonized before building the KG. All these processes are by themselves complex enough and far of being error free therefore the historical data used in this paper has been checked and validated by social science experts to avoid being conditioned by those transcription errors in our experiments.

The next step is to build the KG. We modified the EAR representation by adding relations between attribute values, we named it {\bf Evolution knowledge graph} (EKG). These relations have been used to learn attribute embeddings and attribute value embeddings in a first step and no other embeddings have been learned. Finally, we have learned the attribute weights to be used later during the record linkage process.

At record linkage time, we have considered pairs of candidate records, which can be from different census years, as for instance entities $h$ and $t$ in \figurename~\ref{fig:method}. For all the attributes in common between these two records, we have compared their values by using the same $\EA$ method used on training. Then, the values have been aggregated and weighted accordingly to provide a score between 0 and 1 that gives the matching probability. 

In what follows, we provide the details of the WERL method. We start by introducing the notation used to describe all the different steps to be done for training and records linkage prediction.

\subsection{Evolution knowledge graph}
\label{sec:ekg}

\begin{figure}[t]
    \centering
    \input{KG_representation.tex}
    \caption{Example of a KG representation. RT is the set of relational triples: $(e_1,e_3,r_2)$, $A$ is the attribute set (gray dashed arrows), $V$ is the set of attribute values: $\{v_1,\ldots,v_6\}$ and ET the set of evolution triples: $(v_2,v_3,a_i)$. In this representation will be a missing value between $e_3$ and $e_4$. Top left: KG adjacency matrix.}
    \label{fig:KG}
\end{figure}
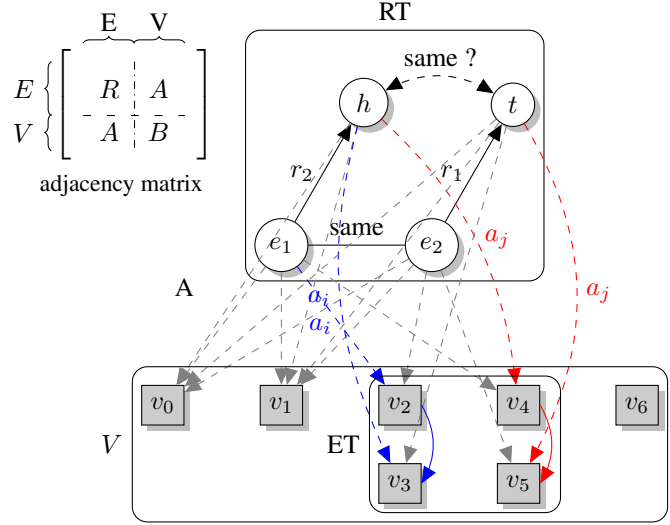

Let $G=(E,V,R,A,B)$ a KG. $E$ and $V$ are the graph nodes while $R$, $A$ and $B$ are the graph edges, see the top left matrix in~\figurename~\ref{fig:KG}.  $E$ is the entity set and $R$ the set of edges between entities. Given two entities $h$ and $t$, respectively called head and tail, a relational triple (RT) is the triple: $(h,t,r)$.  

Regarding KG attributes,  $V$ is the set of the attribute values while $A$, which is called {\em attribute}  set, is the edge set linking entities and attribute values. We denote by AT the set of attribute triples: $(h,v,a)$ and  by $V_a$ the attribute value subset, which is the domain of an attribute $a$. Similarly, $V_e$ is the attribute value set  corresponding to entity $e$. In other words and in the context of census records, $V_e$ corresponds to the actual contents of a record $e$. Although the elements of $V_e$ are  function of entity attributes $a$: $v=v(a)$, we will omit them  when it be clear. Moreover, $A_e$ is  the attributes set of $e$ and   $A_{h,t}=A_h \cap A_t $ the set of attributes shared by both entities $h$ and $t$.

Finally and as being one of our contributions, we add the dependencies between attribute values in our KG representation. These dependencies model the possible changes on  people records through time. To illustrate it, let us consider a kid born in 1867. In first records, he will appear as the son of the head of the household and, probably, his civil status, if reported, single. Later, some years later he would appear as being the head of his own household and probably married. Once he gets married, he would never became single again but could be widow and married. The rationale of adding these attribute values dependencies is to model these semantic dependencies.


Given 2 attribute values $v_i,v_j \in V_a$, that are related because they appear in two linked records, we define the evolution triple (ET) by the triple: ($v_i,v_j,a)$. These triples add arrows between attribute values into the KG, see  \figurename~\ref{fig:KG} for graphical representation of our KG. 

\subsection{Record linkage prediction}

As we have explained at the beginning of this section, we perform record linkage by comparing the attribute values of two candidate records. Given two entities, $h$ and $t$, we take their attribute value embedding vectors, $\theta_{h,a}$ and $\theta_{t,a}$ for all attributes $a\in A_{h,t}$ and we compare them by means of the $\EA$ method used to learn them. Then, we compute the weighted average given by $g$: %
\begin{equation}
g(h,t;\Theta,W) = \sum_{a\in A_{h,t}}w_aS(v,u)\EA(v,u,a;\Theta)
\end{equation}

\noindent  where $v\in V_h$ and $u\in V_t$, $W=(w_a)_a$ is a weight vector associated to attributes $a$, S(u,v) is  set to 0 if $v$ equals $u$ and 1, otherwise,  and $\Theta$ are the embedding vectors learned. To convert the score return by $g$ to a value to be probability interpreted, we apply the sigmoid function,  $\sigma$: %
\begin{equation}\label{eq:rl}
    P(y=1|h,t;\Theta,W) = \frac{1}{1+\exp\{- g(h,t;\Theta,W\}}
\end{equation}

\subsection{First step: embedding learning}

The first step consist of learning the embeddings used to predict record links. To learn them, we can use any of the methods reviewed in Section~\ref{sec:kg_methods} but applied only to the ET set. We do not need to learn embeddings to RT and AT sets since they are not used for record linkage.

Consequently, negative sampling must be done for ET sets. To do so, we generate them  as it is done in similar approaches~\cite{Lin2015PTransE,Lin2016}. For each attribute value $v_i\in V_a$, we sample attribute values from $V_a\setminus E_{v_i}$, where $E_{v_i}=\{ v_j \in V_a \,|\, (v_i,v_j,a)\in ET  \}$. We denote by $E^{+}$ and $E^{-}$ the positive and negative evolution triple samples, respectively. 

We use the same loss function for attribute and attribute values embeddings as the original KG-based embedding method. If we named that loss function $L_{KG}$, the actual loss function used to train the embedding is: %
\begin{align}
L_{EA}(\Theta) = \!\!\sum_{(v,u,a)\in E^{+}} \!\! L_{KG}(v,u,a;\Theta) + \!\! \sum_{(v,u,a)\in E^{-}} \!\! L_{KG}(v,u,a;\Theta)\label{eq:LEA}    
\end{align}

\begin{table*}[ht]
\setlength{\tabcolsep}{5pt} 
    \centering
    \begin{tabular}{cV{2.5}rrrV{2.5}rrrV{2.5}rrr} \toprule
{\bf Data set}    & \multicolumn{3}{cV{2.5}}{ \bf BALL} & \multicolumn{3}{cV{2.5}}{\bf Febrl} & \multicolumn{3}{c}{\bf Cora} \\ \hline
      {\bf  Set }   & {\bf \# Train } & {\bf \# Val. }  & {\bf \# Test}  &  {\bf \# Train } & {\bf \# Val. }  & {\bf \# Test}  & {\bf \# Train } & {\bf \# Val. }  & {\bf \# Test}   \\ \midrule
      Dataset A  & 9,510 & 9,384 & 10,629  & 3,000 & 1,500 & 500  & 470 & 314 & 156   	 \\ 
      Dataset B & 10,629 & 10,722 & 10,536  & 3,000 & 1,500 & 500  & 470 & 313 & 156    \\ 
      Candidate Pairs & 238,854 & 240,954 & 244,696 & 29,066 & 7,674 & 959 & 220,900 & 98,282 & 24,336 \\ 
      True Pairs  & 6,748 & 9,049 & 8,412  & 1,952 & 995 & 340 & 7,961 & 3,601 & 883 \\ 
       \bottomrule
    \end{tabular}
    \caption{Data Partition into train, test and validation sets.}
    \label{tab:data_partition}
\end{table*}

\subsection{Second step: weight learning}

Once we learned the embeddings, we must learn the attribute weights, which provide the impact of each attribute to record linkage prediction. Positive and negative RT, named respectively $T^+$ and $T^-$, come from the candidate pair set. Similarly to the ET set, $T^+$ is composed of all the linked records in the training set while $T^-$ is composed of all the possible linked records that are not in $T^+$. 

The loss function is the hinge loss evaluated on the predicted record linkage probability given by Eq.~\eqref{eq:rl}. This loss is evaluated in both positive and negative tuples as shown below: %
%
\begin{align}
    \begin{split}
    L_{RL}(W) &= \sum_{(h,t)\in T^{+}} \max\left\{0,\lambda_{RL} +  P(y=1|h,t;\Theta,W) \right\} +\\
    &+ \sum_{(h,t)\in T^{-}} \max\left\{0,\lambda_{RL} - P(y=1|h,t;\Theta,W) \right\}\label{eq:LRL}    
    \end{split}
\end{align}

\section{Experiments}

We carried out two experiments in order to evaluate the WERL method. In order to properly evaluate the contributions of this paper, we slightly modified the WERL method. We named MERL the modified method and it consist of not considering weights when predicting record linkage.

The first experiment consists of comparing the WERL performance with some state of the art EA methods, described in Section~\ref{sec:kg_methods}, over  three benchmark data sets detailed in next section, namely BALL, Febrl and Cora data sets.  The second experiment aims at evaluating WERL, and MERL, robustness to data changes. To this end, we  used a model trained on a census coming from one town, and we applied to census from  neighbouring villages.


\subsection{Data sets}

The Baix Llobregat (BALL) Demographic Database provided the census records for \textit{Sant Feliu} town of Barcelona collected in 16 different census from 1828 to 1940\cite{Pujades2019}. We note that the population of the town doubled two decades (1920-1940). The data set contains around 60K records of individuals with 30 attributes. Available attributes include individual's full name, year of birth, civil status, occupation and relationship with head of the family. 


The Freely Extensible Bio-medical Record Linkage (Febrl) is a synthetic data set about bio-medical records of patients~\cite{christen2008}. Each record provides information like full name, address, postal code and birth date. It contains 10K records split across two data sets of 5K records each. For each record, there is a duplicate record in the other data set. We attempt to link these duplicate records across these two data sets.
    
Cora data set contains bibliographical information about scientific papers in XML format~\cite{draisbach2010}. It contains 1879 records of scientific citations with details like title, author(s), publisher, journal and date. The citations refer to 191 unique papers. We attempt to link all citations referring to a same scientific paper. 

We split data into training, validation and test sets. We further partition each set in two data sets $A$ and $B$ to build candidate pairs ($a$, $b$). To reduce the number of candidate pairs in the BALL data set we filter those pairs having the same second surname, see Table~\ref{tab:data_partition} for the counts of each of these sets. Then, we built the corresponding EKG for each of these data sets, see Table~\ref{tab:GraphEvolveKG} for the complete counts.

\begin{table}[ht]
\setlength{\tabcolsep}{4pt} 
    \centering
    \begin{tabular}{|c|c|c|c|c|c|c|c|} \hline
        & {\bf $\#E$} &{\bf $\#A$} & {\bf  $\#R$} & {\bf $\#V$} & {\bf $\#AT$} & {\bf $\#RT$} & {\bf $\#ET$}  \\ \hline
         BALL & 16883 & 6 & 51 & 3492 & 82500 & 13750 & 10643 \\ \hline
         FEBRL & 6000 & 5 & 1 & 7497 & 30000 & 0 & 1112\\ \hline
         Cora & 940 & 15 & 1 & 1764 & 7550 & 0 & 26123\\ \hline
    \end{tabular}
    \caption{Training set sizes of Evolution KG.}
    \label{tab:GraphEvolveKG}
\end{table}

\begin{table*}[ht]
\setlength{\tabcolsep}{5pt} 
    \centering
    \begin{tabular}{cV{2.5}ccccV{2.5}ccccV{2.5}cccc} \toprule
{\bf Data set}    & \multicolumn{4}{cV{2.5}}{ \bf BALL} & \multicolumn{4}{cV{2.5}}{\bf Febrl} & \multicolumn{4}{c}{\bf Cora} \\ \hline
      {\bf  Method }   & {\bf Acc. } & {\bf P }  & {\bf R} & {\bf F-Score} & {\bf Acc. } & {\bf P }  & {\bf R} & {\bf F-Score} & {\bf Acc. } & {\bf P }  & {\bf R} & {\bf F-Score}  \\ \midrule
      TransE (ER) & 0.87 & 0.19 & 0.25 & 0.21 & 0.95 & 0.70 & 0.49 & 0.58 & 0.99 & 0.62 & 0.48 & 0.54 \\ 
      TransH (ER)  & 0.81 & 0.12 & 0.26 & 0.17 & 0.95 & 0.72 & 0.52 & 0.60 & 0.95 & 0.39 & 0.51 & 0.44\\ 
      KR-EAR (EAR) & 0.79 & 0.13 & 0.30 & 0.18 & 0.42 & 0.07 & 0.65 & 0.13 & 0.88 & 0.20 & 0.81 & 0.32 \\ 
      SEEA (EAR) & 0.91 & 0.07 & 0.10 & 0.08	& 0.91 & 0.44 & 0.57 & 0.49 &  0.97 & 0.79 & 0.21 & 0.33\\ \hline
      MERL (ER) & 0.98 & 0.90 & 0.68 & 0.77 & 0.96 & 0.79 & 0.92 & 0.85  & 0.91 & 0.23 & 0.58 & 0.33 \\ 
      WERL (ER) & 0.99 & 0.99 & 0.67 & 0.80 & 0.95 & 0.90 & 0.71 & 0.79  & 0.91 & 0.23 & 0.58 & 0.33 \\ 
      MERL (EKG) & 0.99 & 0.76 & 0.88 & 0.82 & 0.99 & 0.99 & 0.97 & 0.98  & 0.95 & 0.32 & 0.39 & 0.35 \\ 
      WERL (EKG) & 0.99 & 0.98 & 0.89 & 0.93 & 0.97 & 0.99 & 0.76 & 0.86  & 0.97 & 0.61 & 0.36 & 0.46 \\  
    \bottomrule
    \end{tabular}
    \caption{Record Linkage Results for three data sets, presents Accuracy, Precision, Recall and F-score.}
    \label{tab:resultsLinkage}
\end{table*}

\subsection{Methods and parameters}

We implemented various translation based algorithms for generating knowledge graph embedding. We used tensorflow\footnote{ \url{https://www.tensorflow.org/install}} library to generate and optimize embedding for the KG structures defined above. We referred an open-source collection of graph embedding methods called OpenKE\footnote{ \url{https://github.com/thunlp/OpenKE}}, to implement the following  KG embedding based methods for entity alignment: 
\begin{itemize}
    \item TransE and TransH for ER KG
    \item KR-EAR and SEEA for Evolution KG
\end{itemize}
   
To properly evaluate the contribution of each of the elements introduced in the WERL method we have implemented a one more method called: {\bf MERL }. For MERL, we explicitly consider weights as 1 i.e. we consider the mean distance instead of weighted distance. Moreover, we applied these two methods on two KGs: a basic ER knowledge graph, in which attribute are also considered as being entities, and to the EKG introduced in Section~\ref{sec:ekg}. Finally,  we applied grid-search on each of the considered methods to find the optimal hyper-parameters, see Section 1 of the Supplementary Material, for fair comparison.

\subsection{1st experiment. Record linkage  evaluation.}


Overall, EA methods fails when they are applied to record linkage tasks while record linkage methods achieve significantly better performance in all data set in terms of Precision, Recall and F-score, metrics, see Table~\ref{tab:resultsLinkage}. However, in terms of accuracy it seems that all methods obtain relative good results although  WERL and MERL based method still achieve the best results. It should be said that these accuracy values can be misleading since we are in a two highly unbalanced classification problem, see Table~\ref{tab:resultsLinkage}. We note that the major concern is False Positives, since incorrect linking can have a chain effect and have negative impact on the concerned research. Since the proportion of true pairs is very low (3\% census, 7\% febrl, 4\% cora) compared to candidate pairs, it is hard to avoid false positives without increasing false negatives. 


Regarding the BALL data set, the WERL (EKG) method provides the best F-Score of 0.93. Among the compared EA methods, TransE has the highest F-Score of 0.21 only. TransH and KR-EAR has F-Scores of 0.17 and 0.18. We note that we get better results when training over evolution triples. 
%
Concerning the Febrl data set, the MERL EKG provides the best F-score of 0.98. Among EA methods, TransH has the highest F-Score of 0.60, while TransE has a F-Score of 0.58. Applying self-learning to KR-EAR i.e. using SEEA method improves the performance significantly. We note that MERL EKG method perform better than WERL EKG. %
%
Finally, as it regards the CORA data set, %
TransE provides the best F-Score of 0.54. SEEA was able to improve the results slightly from KR-EAR. Among our proposed methods, WERL-EKG provides the highest F-Score of 0.46. We further note that WERL does not perform well on sparse dataset.

In summary, WERL EKG and MERL EKG methods provide satisfactory and stimulating  results since with these methods we are able to learn proper embedding vectors for KGs.

\begin{table}[ht]
    \centering
    \begin{tabular}{|c|c|c|c|c|} \hline
          & {\bf Accuracy } & {\bf Precision } & {\bf Recall} & {\bf  F-Score}  \\ \hline 
          WERL-ET & 0.96 & 0.87 & 0.51 & 0.64 \\ \hline
          MERL-ET & 0.94 & 0.58 & 0.77 & 0.66 \\ \hline
    \end{tabular}
    \caption{Additional Results for Santa Coloma and Castellvi towns}
    \label{tab:additonal_result}
\end{table}

\subsection{2nd experiment. Method robustness.}

Additionally, we tested WERL-ET method on one more census dataset from \emph{Santa Coloma de Cervello} and \emph{Castellvi de Rosanes} towns of Barcelona. The dataset had 8631 records with similar attributes as were provided for \emph{Sant Feliu} town. 

We partition the dataset as Dataset $A$ and $B$ with 4815 and 3816 records. We considered records from census years 1866, 1924, 1936 and 1945 as dataset $A$. For dataset $B$, we selected the records from census years 1901, 1930, 1940 and 1950. We apply blocking indexing on the second surname field to yield 83264 candidate pairs, out of which only 6407 are true links.
The results are provided in Table \ref{tab:additonal_result}. MERL-ET provides the best F-score of 0.66.

\subsection{Discussion}

We analyze the errors in classification for the BALL data set for WERL EKG method. We noticed 1243 false negatives and 96 false positives. Following are the different reasons for the incorrect classification:

\textbf{Full name:} Over 90\% of false positives have different full names while over 60\% false negatives have same full names.



\textbf{Year of Birth:} Over 98\% false negatives have same years of births. All 96 false positives also have same years of birth. Additionally, we had 45 false positives where year of birth is missing for both records.


\textbf{Civil Status:} Over 80\% false negatives have the same civil status. We had 38 false positives with same civil status i.e. about 58\% false positives have different civil status.


\textbf{Occupation:} Over 50\% false negatives had same occupation across years. We had 24 false positives with same occupation i.e. over 70\% false positives have different occupations.


\textbf{Relation:} About 65\% false negatives have same relationship with head. We had 5 false positives with same relationship i.e. about 95\% false positives have different relationship across years.

To sum up, we note that WERL EKG linked the record pairs having valid evolution of attribute values while rejecting the record pairs having invalid evolution, even though they had some common attribute values. Next improvements must overcome these problems.

\section{Conclusions and future work}

In this paper we have proposed a KG based method for record linkage tasks, named WERL. We have enriched knowledge representation by introducing {\bf evolution} relations. These new edges in the knowledge graph encode valid variations (evolution) on individual data. Thanks to this knowledge representation the WERL method is able to learn embedding vectors which better encode census data. Then in the second step, attribute weights are optimized for the record linkage task. This allow to identify which attributes are more discriminant for record linkage tasks. Thereby, we have taken advantage of knowledge graph representation and the associated entity alignment methods and, at the same time, we have trained a specialized method for record linkage tasks.

We have evaluated the proposed method on benchmark data sets and we have compared it to related methods. The achieved results are satisfactory and stimulating since there is still room for improvement. The proposed method must be applied to larger and more heterogeneous census data sets to evaluate both how it scales when data sizes increase and its robustness to data variability. We are quite confident in knowledge graph methods will behave better than existing approaches.



\section*{Acknowledgments}
This work is partially funded by RTI2018-095645-B-C21, Generalitat de Catalunya,  2017 SGR 1783 and the Recercaixa project Xarxes. The Titan V used for this research was donated by the NVIDIA Corporation.

\bibliographystyle{plain}
\bibliography{refs}

\end{document}

%% file: pipeline.tex
    \begin{tikzpicture}[node distance=1.5cm,
    attstyle/.style={draw,fill=gray!40,minimum size=15,drop shadow},
    entitystyle/.style={shape=circle,draw,minimum size=15,fill=white,drop shadow},
    arrowAttstyle/.style={dashed,gray},
    arrowRelstyle/.style={},
    style1/.style={
      matrix of math nodes,
      every node/.append style={text width=#1,align=center,minimum height=2mm},
      nodes in empty cells,
      left delimiter=[,
      right delimiter=],
      }
  ]
    
    \node[entitystyle] (A) at (7,0) {$e_1$};
    \node[entitystyle,right of=A] (B) {$e_2$};
    \node[entitystyle,above right of=A] (C)  {$h$};
    \node[entitystyle,above right of=B] (D)  {$t$};
    
    \path[arrowRelstyle,-] (A) edge  (B);
    \path[arrowRelstyle,dashed,latex-latex] (C) edge[bend left] node[above] (s) {?} (D);
    \path[arrowRelstyle,-latex] (B) edge (D);
    \path[arrowRelstyle,-latex] (A) edge  (C);
     
    \node[draw,dashed,rounded corners=2mm,fit=(A) (B) (C) (D) (s)] (rt) {};
        \node at ($ (rt.north west) + (-45:.5) $) (lrt) {RT};

    \node [attstyle] at ($ (rt.south) + (-110:1.5) $) (v2)  {$v_2$};
    \node [attstyle] at ($ (rt.south) + (-70:1.5) $) (v4)  {$v_4$};

    \node [attstyle,below=5mm of v2]  (v3)  {$v_3$};
    \node [attstyle,below=5mm of v4]  (v5)  {$v_5$};

     \path [arrowAttstyle,-latex] (A) edge (v4);
     
     \path [arrowAttstyle,-latex] (B) edge node[left] {} (v2);
     \path [arrowAttstyle,-latex] (B) edge node[left] {} (v5);
     
     \path [arrowAttstyle,-latex] (C) edge  (v3);
     \path [arrowAttstyle,-latex] (C) edge (v4);
     
     \path [arrowAttstyle,-latex] (D) edge node[left] {} (v3);
     \path [arrowAttstyle,-latex] (D) edge (v5);
     
     \path[-latex] (v2.east) edge[bend left] node[right] {} (v3.east);
     \path[-latex] (v4.east) edge[bend left] node[left] (b2) {} (v5.east);
     
    \node[draw,dashed,rounded corners=2mm,fit=(v2) (v3) (v4) (v5) (b2)] (et) {};
    \node at (lrt |- v2) (let) {ET};

    \node[draw,rounded corners=2mm,fit=(rt) (et),inner sep=2mm,minimum height=5.5cm] (KG) {};
    \node[above=2mm of KG,align=center] (lkg) {\bf \footnotesize Evolution knowledge graph};
    
    \node[text width=3cm,align=center,anchor=north] at ($(lkg.north) +(-8,0)$) (lc) {\bf \footnotesize  Census document \\[-1mm] images};
    \node  at (lc |- rt) (im1) {\includegraphics[width=1.7cm]{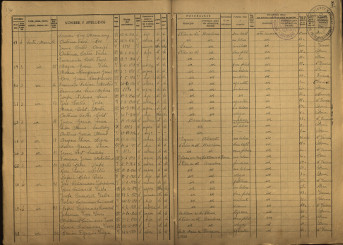}};
    \node[below right=5mm of im1.north west,anchor=north west,label=below:\bf \footnotesize 1940] (im2) {\includegraphics[width=1.7cm]{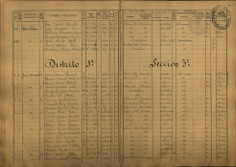}};
     
    \node[below=1cm  of im1]  (im3) {\includegraphics[width=1.7cm]{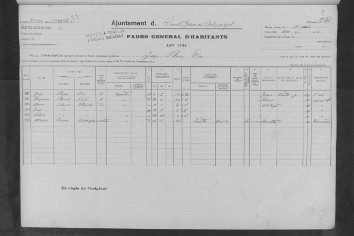}};
    \draw
        node[below right=5mm of im3.north west,anchor=north west] (im4) {\includegraphics[width=1.7cm]{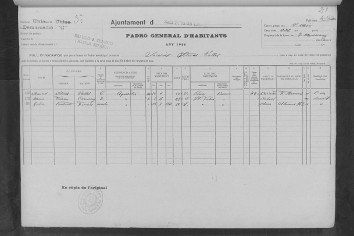}}
        node[below=1mm of im4] (1936) {\bf \footnotesize 1936};

    \node[rounded corners=5mm,draw,fit=(im1) (im4) (1936),inner sep=2mm] (census) {};

    \node[text width=3cm,align=center,anchor=north] at ($ (lc.north -| census.east)!.5!(lkg.north -| KG.west) $) (lie) {\bf \footnotesize Information extraction};
    
    \node[text width=2.5cm,anchor=north] (ie1) at (im1.north -| lie) {\footnotesize Image processing:\\[2mm]
          \begin{varwidth}{\linewidth}\begin{itemize}[leftmargin=*]\setlength\itemsep{1mm}
                            \item Table detection.
                            \item Text extraction \& Recognition.
                            \end{itemize}\end{varwidth}} ;

    \node[anchor=north,text width=2.5cm] (ie2) at (im3.north -| lie) {\footnotesize Text processing:\\[2mm]
        \begin{varwidth}{\linewidth}\begin{itemize}[leftmargin=*]\setlength\itemsep{1mm}
                            \item Cleaning.
                            \item Standardization \& Harmonization.
                            \end{itemize}\end{varwidth}};
                            
    \node[rounded corners=5mm,draw,fit=(ie1) (ie2),inner sep=2mm] (ie) {};
    

     \matrix[right=of et,style1=0.2cm] (1mat)
                                {
                                  & &  \\
                                  & &  \\
                                  & &  \\
                                  & &  \\
                                  & &  \\
                                  & &  \\
                                  & &  \\
                                  & &  \\
                                  & &  \\
                                  & &  \\
                                };
    \node[below=-1mm of 1mat] (Theta) {$\Theta$};

    \draw[dashed]    (1mat-2-1.south west) -- (1mat-2-3.south east);
    \node at (1mat-2-2) {$\theta_{h,a}$};
    \draw[dashed]    (1mat-2-1.north west) -- (1mat-2-3.north east);
    
    \draw[dashed]    (1mat-5-1.south west) -- (1mat-5-3.south east);
    \node at (1mat-5-2) {$\theta_{t,a}$};
    \draw[dashed]    (1mat-5-1.north west) -- (1mat-5-3.north east);

    \draw[dashed]    (1mat-8-1.south west) -- (1mat-8-3.south east);
    \node at (1mat-8-2) {$\theta_a$};
    \draw[dashed]    (1mat-8-1.north west) -- (1mat-8-3.north east);

   \matrix[right=1 of 1mat,style1=0.1cm] (2mat)
                                {
                                  \\
                                  \\
                                  \\
                                  \\
                                  \\
                                  \\
                                };
  
  \node[] at (Theta -| 2mat) (EA) {$EA$};
      \draw[dashed]    (2mat-2-1.south west) -- (2mat-2-1.south east);
    \draw[dashed]    (2mat-2-1.north west) -- (2mat-2-1.north east);
    
    \draw[dashed]    (2mat-5-1.south west) -- (2mat-5-1.south east);
    \draw[dashed]    (2mat-5-1.north west) -- (2mat-5-1.north east);

   \draw[-latex,blue] plot [smooth] coordinates { (1mat-2-3.east) ($(1mat-2-3.east) +(.3,0) $) ($(2mat-2-1.north west)+(-.3,0)$) (2mat-2-1.west)};
  \draw [-latex,red] (1mat-5-3.east) to [bend left=5] (2mat-2-1.west);

    \draw [-latex] (1mat-8-3.east) to [bend left=10] (2mat-2-1.west);

    \node[entitystyle,position=9:1cm from rt]  (C1)  {$h$};
    \node[entitystyle,position=-9:1cm from rt] (D1)  {$t$};
    \node[rounded corners=2mm,fit=(C1) (D1)] (e1) {};
    \node[circle,draw] (j1) at (e1 -| EA) {$g$};
    \node[right=4mm of j1,circle,draw] (s) {$\sigma$} edge[latex-] (j1) ;
    \coordinate (j0) at ($ (e1 -| 1mat.east) + (.5,0)$);
    
    \draw[blue,-latex] (C1) -| (j0) |- (1mat-2-3.east);
    \draw[red,-latex] (D1) -| ($(j0)+ (-.2,-1) $) |- (1mat-5-3.east);

    \node[circle,draw]  at ($ (j1)   + (0,1)$)  (w) {$W$};

    \draw[-latex] (w) --(j1);
    \draw[-latex] (2mat) --(j1);
    
    \node (l1) at ($ (2mat) +(2.7,0) $) {$L_{EA}(\Theta)$} ;

    \node (l2) at (j1 -| l1)  {$L_{RL}(W)$}; 

    \node[rounded corners=2mm,draw,fit=(l1) (let) (et) (EA) (1mat) (Theta),label=below:{\bf \footnotesize 1st step: embedding learning}]  {};
    
    \node[rounded corners=2mm,draw,fit=(l2) (w) (e1)] (T) {};
    
    \node at (lkg -| T) {\bf \footnotesize 2nd step: weight learning};
    
    
    \node at ($ (census.east)!0.5!(ie.west) $) (ac)  {\tikzfancyarrow[5mm]{}};
    \node at ($ (KG.west)!.5!(ie.east) $)  {\tikzfancyarrow[5mm]{}};
    \node at ($ (et.east)!0.5!(1mat.west) $) (etm) {\tikzfancyarrow[5mm]{}}; 
    
    \node at ($ (rt.east)!0.5!(e1.west) $) (rte) {\tikzfancyarrow[5mm]{}};

    \node at ($ (l2.west)!0.5!(s.east) $) (al2) {\tikzfancyarrow[5mm]{}};
    \node at (al2 |- l1)  {\tikzfancyarrow[5mm]{}};

    \end{tikzpicture}

%% file: KG_representation.tex
    \begin{tikzpicture}[node distance=2cm,
    attstyle/.style={draw,fill=gray!40,minimum size=15,drop shadow},
    entitystyle/.style={shape=circle,draw,minimum size=15,fill=white,drop shadow},
    arrowAttstyle/.style={dashed,gray},
    arrowRelstyle/.style={}]
    
    \node[attstyle] (v0)  {$v_0$};
    \node [attstyle,right=10mm of v0]  (v1)  {$v_1$};
    \node [attstyle,right=10mm of v1]  (v2)  {$v_2$};
    \node [attstyle,right=10mm of v2]  (v4)  {$v_4$};
    \node [attstyle,right=10mm of v4]  (v6)  {$v_6$};
    
    \node [attstyle,below=5mm of v2]  (v3)  {$v_3$};
    \node [attstyle,below=5mm of v4]  (v5)  {$v_5$};

    
    \node[entitystyle,above=1.5 of v1] (A)  {$e_1$};
    \node[entitystyle,right of=A] (B) {$e_2$};
    \node[entitystyle,position=60:1.5 from A] (C)  {$h$};
    \node[entitystyle,position=60:1.5 from B] (D)  {$t$};
    
    \path[arrowRelstyle,-] (A) edge node[above]  {same} (B);
    \path[arrowRelstyle,dashed,triangle 45-triangle 45] (C) edge[bend left] node[above] (s) {same ?} (D);
    
     \path[arrowRelstyle,-triangle 45] (B) edge node[left,midway] {$r_1$} (D);
     \path[arrowRelstyle,-triangle 45] (A) edge node[left,midway] {$r_2$} (C);
     
     \node[draw,rounded corners=2mm,fit=(A) (B) (C) (D) (s),label=RT] (RT) {};

     \path [arrowAttstyle,-triangle 45] (A) edge node[left] {} (v0);
     \path [arrowAttstyle,-triangle 45] (A) edge node[left] {} (v1);
     \path [arrowAttstyle,blue,-triangle 45] (A) edge node[left] {$a_i$} (v2);
     \path [arrowAttstyle,-triangle 45] (A) edge node[left] {} (v4);
     
     \path [arrowAttstyle,-triangle 45] (B) edge node[left] {} (v0);
     \path [arrowAttstyle,-triangle 45] (B) edge node[left] {} (v1);
     \path [arrowAttstyle,-triangle 45] (B) edge node[left] {} (v2);
     \path [arrowAttstyle,-triangle 45] (B) edge node[left] {} (v5);
     
     \path [arrowAttstyle,-triangle 45] (C) edge node[left] {} (v0);
     \path [arrowAttstyle,-triangle 45] (C) edge node[left] {} (v1);
     \path [arrowAttstyle,blue,-triangle 45] (C) edge[bend right=20]  node[left] {$a_i$} (v3);
     \path [arrowAttstyle,red,-triangle 45] (C) edge[bend left=20]  node[midway,right] {$a_j$} (v4);
     
     \path [arrowAttstyle,-triangle 45] (D) edge node[left] {} (v0);
     \path [arrowAttstyle,-triangle 45] (D) edge node[left] {} (v1);
     \path [arrowAttstyle,-triangle 45] (D) edge node[left] {} (v3);
     \path [arrowAttstyle,red,-triangle 45] (D) edge[bend left=30]  node[midway,right] {$a_j$} (v5);
     
     \path[blue,-triangle 45] (v2.east) edge[bend left] node[right] {} (v3.east);
     \path[red,-triangle 45] (v4.east) edge[bend left] node[left] (b2) {} (v5.east);
     
    \node[draw,rounded corners=2mm,fit=(v2) (v3) (v4) (v5) (b2),label=left:ET] (b) {};
        \node[draw,rounded corners=2mm,fit=(v0) (b) (v6),label=left:$V$] (V) {};

     \node   at ($ 0.5*(v0) + 0.5*(A) +(-5mm,5mm) $) (attribute) {A};
     
     \draw
     node[] at ([xshift=-3mm,yshift=-1cm]V.west |- RT.north) (Adj) {\input{adjacency.tex}};
     
    \end{tikzpicture} 